# A Structurally and Temporally Extended Bayesian Belief Network Model: Definitions, Properties, and Modeling Techniques

**Constantin F. Aliferis and Gregory F. Cooper**
Section of Medical Informatics & Intelligent Systems Program
University of Pittsburgh, B50A Lothrop Hall, 190 Lothrop St.
Pittsburgh, PA 15261

## Abstract

We developed the language of Modifiable Temporal Belief Networks (MTBNs) as a structural and temporal extension of Bayesian Belief Networks (BNs) to facilitate normative temporal and causal modeling under uncertainty. In this paper we present definitions of the model, its components, and its fundamental properties. We also discuss how to represent various types of temporal knowledge, with an emphasis on hybrid temporal-explicit time modeling, dynamic structures, avoiding causal temporal inconsistencies, and dealing with models that involve simultaneously actions (decisions) and causal and non-causal associations. We examine the relationships among BNs, Modifiable Belief Networks, and MTBNs with a single temporal granularity, and suggest areas of application suitable to each one of them.

## 1. INTRODUCTION

The Bayesian Belief Network (BN) is a state-of-the art formalism for representing uncertainty in a way that is consistent with the axioms of probability theory [Pearl 1988, Neapolitan 1990]. BNs are graphical representations that allow precise and concise descriptions of probabilistic dependencies and independencies among propositional variables in a problem-solving domain of interest. They have a number of important properties, including the ability to capture any joint probability distribution, direct affinity with decision theory that leads to natural development of normative decision-support systems, and the potential for the developer to sometimes specify large joint probability distributions by specifying a few local prior and conditional probability distributions [Neapolitan 1990]. Moreover, inference has been studied and developed extensively in BNs. There are several exact and approximate inference algorithms for use with BNs, and although general inference has been shown to be NP-hard [Cooper 1990, Dagum et al. 1993B], special-case algorithms and corresponding conditions have been established that allow tractable inference [Neapolitan 1990]. Finally, great attention has been given, and substantial results obtained, in the field of automated learning of BNs from data [Cooper et al. 1992, Spirtes et al. 1992, Heckerman et al. 1994, Bouckaert 1994].

BNs were not designed to model temporal relationships explicitly. In many problem areas, however, the ability to model effectively the temporal aspects of the domain plays a crucial role in the success of the modeling effort. For example, in medicine, representing and reasoning about time is crucial for broad reasoning tasks like prevention, diagnosis, therapeutic management, prognosis, and discovery. Similarly, in areas such as economics, biology, and scheduling (among others), capturing the dynamic aspects of the problem at hand is essential for successful problem solving.

Time modeling is an area that has also been intensively explored in a number of scientific fields that are characterized by vastly different perspectives, such as philosophy, physics, statistics, operations research, and more recently artificial intelligence [van Fraassen 1970, Allen 1984, Tansel et al. 1993, Haddaway 1994]. This research shows that devising computer systems that can utilize the temporal dynamics of a problem area to provide decision support (a major goal of AI) is difficult on at least three levels:
(a) *Temporal expressiveness*: It is difficult to develop computational formalisms (i.e., representations) that allow us to express the time-evolving and sensitive domain knowledge in a natural way for a wide variety of domains or tasks within a domain.
(b) *Temporal knowledge acquisition*: It is hard to find experts and/or data that would allow the instantiation of temporal models with the appropriate knowledge. Sometimes the available knowledge is more abstracted than the model language asks for. Often the expert is not comfortable specifying the full temporal evolution of a particular domain process, although he can partially specify this process in terms of summary (i.e., temporally abstracted) and qualitative temporal relationships.
(c) *Computational tractability*: Temporal models typically are much more detailed than atemporal ones. Even when they involve a few variables, examining these and their interactions over multiple points of time often entails an inordinate amount of computation, due to the size and the complexity of the resulting model.

These problem areas readily suggest that any successful time-modeling formalism should be designed so that it satisfies temporal expressiveness, computational efficiency and temporal knowledge acquisition requirements. We call these properties *temporal desiderata,* since they are so crucial for time modeling. Additional desiderata are ones that are considered useful across all knowledge representations, not just temporal ones (we call those *universal desiderata*). In particular, the ideal representation should handle uncertainty in a principled and unambiguous way. It should support normative decision-making. It should handle causality, causal manipulation, and have a declarative semantics. An ideal representation would also be *sound* and *complete*, as well as amenable to machine-learning and explanation methods.

Standard BNs do satisfy the universal desiderata. They handle uncertainty using the well-defined and extensively studied language of probability theory. BNs support decision-theoretic rationality, a framework for making normative decisions very well suited to many problems, such as economic and medical ones. BNs can be interpreted causally both for discovery and optimal decision-making purposes [Spirtes 1992, Drudzel et al. 1993]. BNs do have declarative semantics and support sound and complete reasoning (in the sense that the correct probability of any proposition of variables can be derived conditioned on any other proposition of variables). Finally, several machine learning and explanation methodologies have been under investigation, with promising results during the last decade [Cooper et al. 1992, Spirtes et al. 1992, Heckerman et al. 1994, Bouckaert 1994, Suermondt et al. 1993].

These positive properties of BNs, suggest building a language of time modeling on them by adding new features that facilitate temporal representation and reasoning. This explains why an important trend in the study and use of BNs is their extension, usage and/or analyses to accommodate decision, causal, temporal and planning models of problem solving [Provan 1993, Drudzel et al. 1993, Ngo et al. 1995]. Some of these extensions are practical approaches that emphasize achieving problem solutions with minimal alterations in the base model [Provan 1993, Polasheck et al. 1993]. Other researchers have extended the language of BNs [Dagum et al. 1993A, Hanks et al. 1995, Darwiche et al. 1994].

In [Aliferis et al. 1995], we introduced Modifiable Temporal Belief Networks (MTBNs), a temporal and structural extension to BNs that model time-sensitive domains in medicine. Although our point of departure is medical, the generality of our approach (as it will be discussed in this paper) should make MTBNs applicable to non-medical domains as well. In this paper, we discuss MTBNs by providing definitions of the model and examining its theoretical properties. We also examine temporal modeling techniques, compare different classes of MTBNs, and discuss conclusions and future research.

## 2. AN EXAMPLE OF AN MTBN MODEL

In this section we give an example of using MTBNs to represent and apply temporal knowledge to solve a small temporal reasoning task. The purpose of the example is to give an intuitive notion of MTBNs, before giving more formal definitions.

Figure 1 presents a small MTBN that captures a fundamental causal mechanism: the feedback between glucose $(G)$ and insulin $(I)$ in the human body. An increased glucose blood level triggers the secretion of insulin, which in turn causes the glucose level to drop. Low levels of glucose cause the secretion of insulin to drop, which in turn might allow the glucose level to rise.

We know that the presence of diabetes mellitus $(DM)$ can interfere with this regulatory mechanism. In this simplistic model, and solely for the purposes of demonstration, assume that $DM$ interferes by modulating the value of the lag (delay) between changes in the insulin and change in the glucose level. In particular, $DM$ is associated with higher chances of a prolonged delay. We emphasize that this example is intended only to show some basic MTBN modeling principles and is not intended to be medically accurate.

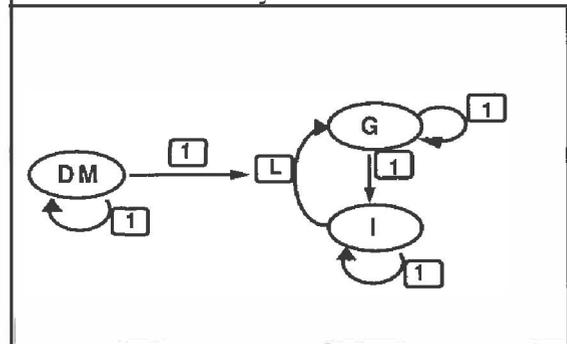

Figure 1: A simple MTBN model of glucose-insulin interaction ($DM$=diabetes mellitus, $G$=glucose, $I$=insulin).

This MTBN model makes it possible to answer questions such as :

- Given that at time point 1, 2, and 3 we know that *diabetes mellitus*=true, and that at time 1 *glucose*=high, what is the probability that *glucose* at time 3 will be high?
- Given that we have measurements of *glucose* and *insulin* at points 1,2,3, what is the probability that *diabetes mellitus*=false?

In table 1 we give a specification for this MTBN, as well as examples of terminology. In the specification below, "$X_t$" means "variable $X$ at time $t$". Qualitatively, we would say for this example that higher glucose causes higher insulin, higher insulin causes lower glucose, and that diabetes causes a more delayed response of glucose levels to insulin levels.

Table 1 : Parameter and variable specifications for the MTBN structure of Figure 1.

---

Temporal range = 1 to 10.
Temporal unit = 10 min.
Variables :
(a) Ordinary: $G$ (1=low, 2=medium, 3=high), $I$ (1=low, 2=medium, 3=high), $DM$ (1=yes, 2=no)
(b) Causal mechanisms: $[G\text{->}I]$, $[I\text{->}G]$, $[DM\text{->}LAG[I\text{->}G]]$, $[G\text{->}G]$, $[I\text{->}I]$, $[DM\text{->}DM]$. All arc variables are activated.
(c) Lag variables: $LAG[G\text{->}I]$=1, $LAG[I\text{->}G]$ =(1 or 2), $LAG[DM\text{->}LAG[I\text{->}G]]$=1, $LAG[G\text{->}G]$=1, $LAG[I\text{->}I]$=1, $LAG[DM\text{->}DM]$=1.
Conditional probability distributions* :
(a) if no information on the previous value of $DM$ is known (i.e., time=1) then:
    $p(DM_t = 1 / 2) = 0.1 / 0.9$
    If time>1 then:
    $p(DM_t = 1 / 2 \mid DM_{t-1} = 1) = 1 / 0$
    $p(DM_t = 1 / 2 \mid DM_{t-1} = 2) = 0.001 / 0.999$
(b) If no information on previous value of I is known (i.e. time=1) then:
    $p(I_t = 1 / 2 / 3 \mid G_t = 1) = 0.7 / 0.2 / 0.1$
    $p(I_t = 1 / 2 / 3 \mid G_t = 2) = 0.2 / 0.6 / 0.2$
    $p(I_t = 1 / 2 / 3 \mid G_t = 3) = 0.1 / 0.2 / 0.7$
    If time>1 then:
    $p(I_t = 1 / 2 / 3 \mid G_{t-1} = 1, I_{t-1} = 1) = 0.8 / 0.15 / 0.05$
    . . .
    $p(I_t = 1 / 2 / 3 \mid G_{t-1} = 3, I_{t-1} = 3) = 0.05 / 0.15 / 0.8$
(c) If no information on previous value of $G$ is known (i.e. time=1) then:
    $p(G_t = 1 / 2 / 3) = 0.1 / 0.8 / 0.1$
    If time>1) and $LAG[I\text{->}G]$ is L, then:
    $p(G_t = 1 / 2 / 3 \mid I_{t-L} = 1, G_{t-1} = 1) = 0.075 / 0.075 / 0.85$
    . . .
    $p(G_t = 1 / 2 / 3 \mid I_{t-L} = 3, G_{t-1} = 3) = 0.65 / 0.25 / 0.10$
(d) If no information on previous value of $DM$ is known (i.e. time=1), then:
    $p(LAG[I\text{->}G]_t = 1 / 2) = 0.9 / 0.2$
    If time>1 then:
    $p(LAG[I\text{->}G]_t = 1 / 2 \mid DM_{t-1} = 1) = 0.2 / 0.8$
    $p(LAG[I\text{->}G]_t = 1 / 2 \mid DM_{t-1} = 2) = 0.99 / 0.01$

---

*The notation: '$p(X=1/, ... , /n)=Val_1/, ... , /Val_x$', means : $p(X=1)=Val_1, ... , p(X=n)=Val_n$.

---

This example informally introduces some important concepts of MTBN models. First MTBNs involve probabilistic causal relations among propositional variables. The variables are examined for a period of time. Variables are measured at discrete time points. The delay between cause and effect is itself a variable which is modeled explicitly. The causal relations between variables are themselves variables, which are modeled explicitly. The model is fully specified by defining how each variable is determined by its parents (given different conditioning of the relevant arc and lag variables). We note that in this example an additional feature of MTBNs is not depicted, specifically the ability to model 'abstract' or non-indexed variables. These will be explained in the subsequent sections.

## 3. DEFINITIONS OF THE MTBN MODEL

We now formalize the components of an MTBN.

**Definition 3.1.** An MTBN-SG (single-granularity MTBN) is defined as MTBN-SG = {Gt, Tr, G, D, J}, where each of the elements is defined as follows.

Gt (temporal graph): This is a directed graph (possibly cyclic) composed of nodes and arcs corresponding to 3 types of variables. The first type of variable is an **ordinary variable,** semantically corresponding to (potentially) observable phenomena. An ordinary variable is either explicitly associated with a time point (i.e., is *temporally indexed*) or not (i.e., is *abstract*). The interpretation of an abstract variable is that it is assigned a value at some time within the temporal range of the model. This time is not specified and can only be constrained by causal associations of the abstract variable with other variables in the model. The set of all ordinary variables is denoted by *V*. All ordinary variables are represented in the graph by a node that has a symbolic name. The second type of variable is a **mechanism variable** (also called an "arc variable"), semantically corresponding to causal mechanisms between variables. The set of all mechanism variables is denoted by *E*. In an MTBN graph, all variables in *E* are represented as an *arc* between two other variables. Mechanism variables take one of two possible values (true or false, that is, active or inactive). As a convention, arc variables that do not have the constant value 'true' are denoted graphically by an arc with a circle. The third type of variable is **time-lag quantifier variable**, semantically corresponding to the time lag (expressed in time units and being positive) between a variable $V_1$ (the cause) and a variable $V_2$ (the effect). The set of time-lag quantifying variables is denoted by *L*. Every variable in *L* is represented in Gt as a square node associated with an arc (mechanism variable). This node has a symbolic name, or in the case it has a constant value, the name can be replaced by a numeric constant. We will use the symbol $X$ to denote a variable of any type. We will also adopt all the standard graph-theoretic notions of parent, child, sibling, descendant (direct, indirect), ancestor (direct, indirect), path (directed or not), cycle (directed or not) [Neapolitan 1990].

Tr: the <u>temporal range</u> (or frame of discernment). Tr = $[t_1,...,t_n]$ , where $t_1$ is the initial time point of interest and $t_n$ is the last time point of interest. We associate variables with a discrete, ordered, linear, unbounded model of time, in which temporal intervals are built from temporal points and only a portion of the temporal line is modeled explicitly during problem solving. Instantaneous influences are also allowed and are modeled in MTBNs, as having time lags of constant value 0.

G: the <u>temporal unit granularity</u> This is the unit of time in our model.

D: a <u>deployment transformation</u>. A temporal graph contains information on causal processes and the way they behave and they evolve over time. This information is represented in a compact form in Gt, and thus it has to be unfolded over time to actually model the domain of interest (and perform inference). We call the compact representation the **condensed graph**, and the unfolded one the **deployed graph**. We call the transformation from condensed to deployed form for single-granularity MTBNs the *standard* deployment transformation (SD), and it operates as follows: *n copies of every variable are created. Each variable copy is time-stamped with an index sequentially, starting with $t_i$ and ending with $t_e$. ⬤rdinary variables designated by the user as "abstracted" are not replicated or stamped. Mechanism ⬤nd lag variables are time-stamped with the time-stamp of the cause variable. No indexed variable copy is created outside the MTBN temporal range.*

We use the notation $[X]$ to denote the set of all variable instances. We also use variable symbols in bold font to denote instantiated variables.

J: a <u>joint probability distribution</u> over the variables in the model.

**Definition 3.2.** An <u>active mechanism</u> at a particular time point $t$ is a mechanism that is assigned the value "true" at time $t$. As an example, the arc $[A->B]$ of figure 2.2 is active at t=1, but inactive at time $t=2$.

**Definition 3.3.** Let the set $S = E \cup L$ be called the set of structural variables. In the example of figure 2.1, $S=$ { $[A->B]_1$, LAG$[A->B]_1$, $[A->B]_2$, LAG$[A->B]_2$ }. Since the lag variable has the constant value '0', we can simplify $S$ to be : { $[A->B]_1$, $[A->B]_2$ }.

**Definition 3.4.** We call the $i$th joint instantiation of the variables in $S$ the structure $S_i$ We call any joint instantiation of the variables in $S$ that correspond to a time point $j$ ($j \in \{t_1,..., t_n\}$) the <u>j-th substructure of a structure</u> $S_i$, which we denote as $S_{ij}$. In the example of figure 2.1, the possible structures are: $S_1=$ { $[A->B]$ = active, $[A->B]_2=$active }, $S_2=$ { $[A->B]_1$ = active, $[A->B]_2=$ inactive }, $S_3=$ { $[A->B]_1$=inactive, $[A->B]_2$=active }, $S_4=$ { $[A->B]_1$=inactive, $[A->B]_2$ = inactive }. The first substructure of $S_2$ is $S_{2,1}$ = { $[A->B]_1$=active }, while the second substructure of $S_2$ is $S_{2,2}$ = { $[A->B]_1$=inactive }.

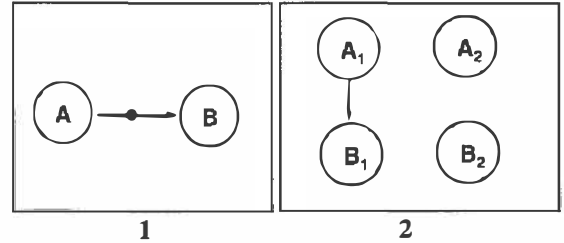

Figure 2: Simple example of MTBN model with one arc variable and two temporal points (2.1.), and one of several possible BN structures for this model (2.2.).

**Definition 3.5. :** $S_i$ is acyclic iff it contains no directed cycles of active mechanisms associated with $S_i$. $S_2$ of figure 2.2 is acyclic.

**Definition 3.6. :** <u>J is well-defined</u> if every cyclic $S_i$ has a probability of 0. The corresponding MTBN-SG is said to be a <u>well-defined MTBN-SG</u>. In the example of figure 2.1, all 4 possible structures are acyclic, thus any probability distribution over the model variables will be well-defined.

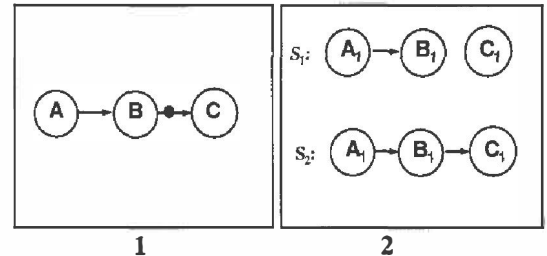

Figure 3: Simple example of MTBN model with two arc variables and a temporal range with just one temporal point (3.1.), and the two possible structures for this model (3.2.).

**Definition 3.7.** Given a structure $S_i$, the <u>active parents of a variable $X_i$ at time $t$</u> (denoted as $X_{i,t}$) are the variables in $S_i$ with active mechanisms into $X_{i,t}$ in $S_i$. Denote the active parents of $X_{i,t}$ as: $c(X_{i,t})_{S_i}$. In the example of figure 2.1., relative to structure $S_1$, the active parent of $B$ at time 1 is $A$ at time 1, and of $B$ at time 2 is $A$ at time 2. Relative to structure $S_2$, the active parent of $B$ at time 1, is $A$ at time 1, while $B$ at time 2 has no active parents.

**Definition 3.8. :** Let an <u>ancestral ordering for a $S_i$</u> be an ordering of the variables $[X]$ such that for every variable $X_j$ at time $t$, no variable to the right of $X_j$ belongs to $c(X_{j,t})_{S_i}$. In the example of figure 3, an ancestral ordering for $S_1$ is: ( $A_1$, $[A->B]_1$, $B_1$, $[B->C]_1$, $C_1$ ), or ( $C_1$, $A_1$, $[A->B]_1$, $B_1$, $[B->C]_1$). An ancestral ordering for $S_2$ is: ( $A_1$, $[A->B]_1$, $B_1$, $[B->C]_1$, $C_1$ ).

**Definition 3.9.** : We denote a constraint on the values of the structural variables of $[X]$ to be those entailed by structure $S_i$ as: $[X]_{S_i}$. Similarly the joint probability of $[X]$ constrained by $S_i$ will be denoted as: $p([X])_{S_i}$. Let a set of constraints on a joint probability distribution J be a <u>structural constraint set on J (SCS(J))</u> iff it is the set of all pairs : ( $[X]_{S_i}$, $p([X])_{S_i}$ ). SCS(J) contains exactly as many members as the number of structures $S_i$. In the example of figure 3: $p(A_1, [A\text{->}B]_1, B_1, [B\text{->}C]_1, C_1)_{S_1} = p(A_1, [A\text{->}B]_1, B_1, [B\text{->}C]_1$=inactive, $C_1)$. In contrast, $p(A_1, [A\text{->}B]_1, B_1, [B\text{->}C]_1, C_1)_{S_2} = p(A_1, [A\text{->}B]_1, B_1, [B\text{->}C]_1$=active, $C_1)$.

**Axiom 3.1.** (<u>MTBN-SG Markov Condition</u>): The probability distribution of any variable $X_{it}$ at time $t$, given its active parent variables, is independent of any subset $W$ of the non-descendent nodes of $X_{it}$ for each structure $S_i$. That is, for each $S_i$: $p( X_{it} ) \mid c(X_{it})_{S_i} \cup W) = p( X_{it} ) \mid c(X_{it})_{S_i}$. In the example of figure 3, relative to $S_1$ : $p(C_1 \mid B_1, A_1) = p(C_1)$, but relative to $S_2$: $p(C_1 \mid B_1, A_1) = p(C_1 \mid B_1)$.

## 4. PROPERTIES OF THE MTBN-SG MODEL

In this section, we describe a number of fundamental MTBN-SG properties which have obvious parallels to properties of BNs. Detailed derivations can be found in [Aliferis et al. 1996]. Here we provide the sketches of proofs only.

We first show that variables in a deployed MTBN-SG can be ordered according to their causal relations, and the particular instantiations of the structural variables. This will be important for both subsequent properties and inference.

**Lemma 4.1.** (<u>Ancestral ordering result</u>). Let $M_1$ be an arbitrary well-defined MTBN-SG. The structural constraint set on J determines a set of ancestral orderings on $[X]$, s.t. for each possible structure $S_i$ there exists at least one ancestral ordering ord($S_i$) of $[X]$.
**Proof (sketch)**: The following procedure produces ancestral orderings for each $S_i$. Take all variable instances over time, with no active parents and put them first in the ordering (ties are broken arbitrarily). Find the variables with all their active parents already ordered and order them subsequently (again ties are broken arbitrarily). Repeat until no more variables are unordered. It can be shown that this procedure terminates, that upon termination no variables are left unordered and that when it terminates no ordered variable comes before any of its active parents (i.e., we have a valid ancestral ordering) .                 □

In the example of figure 3, J can be decomposed as follows: $\{p(X)_{S_1}, p(X)_{S_2}\}$. For each one of $S_1$ and $S_2$ we can obtain ancestral orderings. For example, ord($S_1$) = $(A_1, [A\text{->}B]_1, B_1, [B\text{->}C]_1, C_1)$.

Now we show that for each possible joint instantiation of the structural variables, we can factorize the joint probability distribution according to the prior and conditional distributions of each variable given its active parents (if any). This result is important in terms of proving subsequent properties, establishing a closer correspondence of BNs and MTBNs, as well as for developing inference algorithms.

**Lemma 4.2.** (<u>MTBN factorization result</u>). For a joint probability distribution J of an arbitrary MTBN-SG $M_1$, there exists a factorization of J for each $S_i$.
**Proof (sketch)**: By Lemma 4.1, for each $S_i$ of $M_1$ there exists an ancestral ordering ord($S_i$) on $[X]$. It suffices to show that for an arbitrary ancestral ordering $S_i$ we can derive a factorization of the partition of J. Let the ordered set of variables according to ord($S_i$) be: $\{ X_1,..., X_q \}$, s.t. all active parents of $X_i$ are to the left of $X_i$. Then by applying the chain rule of probabilities, we have:
$p(X_1,..., X_q) = p (X_q \mid X_{q-1},..,X_1) \, p (X_{q-1} \mid X_{q-2},..,X_1),..., p (X_1)$

Using axiom 3.1 and the fact that we have an ancestral ordering, we get:    $p(X_1,..X_q) = p(X_q \mid c(X_q) )_{S_i} \, p(X_{q-1} \mid c(X_{q-1}) )_{S_i},...,p(X_1)_{S_i}$, or equivalently,

$$p (X)_{S_i} = \prod_j p(X_j \mid c(X_j)_{S_i})$$

(The $j$ index enumerates all variables in the model at all time points in the temporal range).                 □

Thus, for each possible structure $S_i$ we have a factorization corresponding to the respective structural constraint on the joint probability distribution J of $M_1$.

In the example of figure 3, $p(A_1, [A\text{->}B]_1, B_1, [B\text{->}C]_1, C_1)_{S_1} = p(A_1) * p( [A\text{->}B]_1 = $ active$) * p( B_1 \mid A_1 ) * p( [B\text{->}C]_1$=inactive $) * p( C_1)$, and $p(A_1, [A\text{->}B]_1, B_1, [B\text{->}C]_1, C_1)_{S_2} = p(A_1) * p( [A\text{->}B]_1 = $ active$) * p( B_1 \mid A_1 ) * p( [B\text{->}C]_1 = $ active $) * p( C_1 \mid B_1)$.

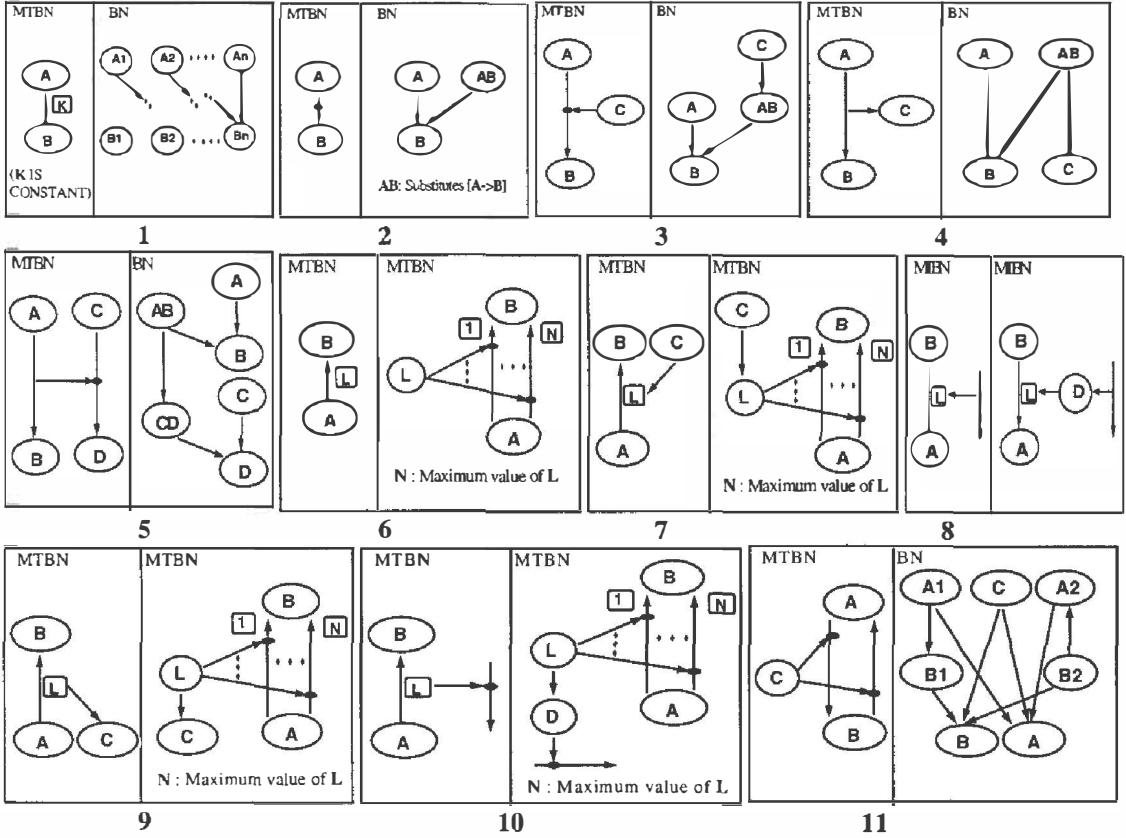

Figure 4: Graphical transformations of MTBN-SG to BN format.

**Corollary 4.1.** (Compactness result) We can retrieve the joint probability distribution J of an arbitrary MTBN-SG from the conditional probability distributions $p(X_j \mid c(X_j)_{s_0})$.
**Proof** : From Lemma 4.2 and by marginalizing over all possible instantiations of $S$ we get:

$$p(X) = \sum_{S_i} p(X)_{g_i} = \sum_{S_i} \prod_j [p(X_j \mid c(X_j)_{s_0})]$$

(where $S_i$ indexes all possible structures).  □

We note that no term $P(S_i)$ is needed in the equation since $S_i$ itself contains structural variable value assignments that represent the probability of structure $S_i$. Not only do we do not need to include the prior probabilities of possible structures in the equation of Lemma 4.2., but doing so would not allow us to do computation with MTBNs, since we cannot in general estimate the prior of any structure without doing inference on the net. This is because in MTBNs structural variables may be determined by ordinary and structural variables. The only case where such an approach would work is when dynamic structural variables would only have prior distributions (no parents). In the example of figure 3, $p(A_1, B_1{=}\text{false}, C_1, [A{-}{>}B]_1, [B{-}{>}C]_1 ) = p(A_1, B_1{=}\text{false}, C_1, [A{-}{>}B]_1, [B{-}{>}C]_1 )_{S1} + p(A_1, B_1{=}\text{false}, C_1, [A{-}{>}B]_1, [B{-}{>}C]_1 )_{S2} = p(A_1) *$

$p([A{-}{>}B]_1 = \text{active}) * p(B_1{=}\text{false} \mid A_1 ) * p([B{-}{>}C]_1{=}\text{inactive} ) * p(C_1) + p(A_1) * p([A{-}{>}B]_1 = \text{active}) * p(B_1 = \text{false} \mid A_1 ) * p([B{-}{>}C]_1{=}\text{active}) * p(C_1 \mid B_1{=}\text{false}).$

Since a BN can capture any joint probability distribution, we can always construct a BN that captures the joint probability distribution J of an arbitrary MTBN. We show next that we can also always develop a BN that captures J, while preserving semantic equivalence of the variables in the two models.

**Theorem 4.1.** (MTBN transformability to BNs result). An arbitrary MTBN-SG $M_1$ can be transformed to a standard BN $B_1$, s.t. $B_1$ captures the same joint probability distribution as $M_1$ and each variable in $M_1$ has a (semantically) corresponding variable (or set of variables) in $B_1$.
**Proof (outline)** We employ ten MTBNs-SG to BN, and MTBN-SG to MTBN-SG transformations (depicted in figures 4.1 to 4.10). From Definition 3.1., each transformation captures an essential feature of MTBN-SGs that distinguishes them from BNs. We prove transformability for each MTBN. To do so we utilize the following technique: The original MTBN by definition contains a joint probability distribution J. J can be factored according to Lemma 4.2. We construct a BN by defining conditional probability distributions corresponding to the factorization of the MTBN. By BN properties we know that we thus define a valid BN capturing a unique joint probability distribution. Moreover the two distributions are identical since they are factored in the same fashion. In the complete proof we prove that: (a) each of the transformations of figures 4.1 to 4.10 follow from Definition 3.1. (consistency), (b) each of the transformations yield a valid BN or MTBN (validity), and (c) that any MTBN can be completely transformed to a BN (completeness). In the remainder of the proof we generalize to the case of MTBN-SGs employing all the features together, and dealing with cyclic structures (see figure 4.11). In all cases, we assume $Tr = [t_1,...,t_n]$, and that we introduce in the BN $n$ copies of each variable indexed by $j$ s.t. $j \in [t_1,...,t_n]$.    $\square$

From thm 4.1. and the BN properties [Neapolitan 1990], we immediately get the following corollaries:

**Corollary 4.2.** (Specification result). If we specify a temporal graph Gt and conditional probability distributions for all the variables, then we have defined a unique joint probability distribution and corresponding MTBN-SG. In this paper we do not discuss details on how to specify generalized temporal conditional probabilities (i.e., when for every time point a variable is determined in the same way by its parents at that point). These can be found in [Aliferis et al. 1996]. There we also discuss how to enforce the requirement for a well-defined MTBN-SG when defining a model.

**Corollary 4.3.** (Expressiveness result). For every joint probability distribution J, there exists a MTBN-SG that can capture it.

**Corollary 4.4.** (Inference result). Inference with MTBN-SGs can be carried out with any standard BN inference algorithm by transforming an MTBN-SG to a BN.

**Corollary 4.5.** (Complexity result) Inference with MTBN-SGs (exact and approximate) is NP-hard in the general case. We note here that we have developed temporal versions of Logic Sampling and Likelihood Weighting for doing inference with MTBN-SGs. We have implemented a program called HARMONY that does case simulation and inference with MTBN-SG models. Our program allows causal manipulations and takes advantage of arc deactivation to avoid unnecessary computation [Aliferis et al. 1996].

## 5. MODELING TECHNIQUES

MTBNs can help in the modeling of a wide variety of temporal and causal knowledge. We present several simplified example application vignettes in sections 5.1. and 5.2. In section 5.1 we discuss modeling techniques that are fairly standard with ordinary BNs. In section 5.2. we describe modeling that is particularly facilitated by the MTBN-SG model's special features.

## 5.1. NON MTBN-SPECIFIC MODELING TECHNIQUES

### 5.1.1. Representing intervals and duration of intervals

An interval in an MTBN is represented via two random variables *INT_START* and *INT_END*, which correspond to the endpoints of the interval. Since the start and end of the interval are random variables, we have to ensure that the actual (i.e., instantiated) start of an interval will always be before its actual end. This constraint is enforced with an arc from *INT_START* to *INT_END* that has an associated conditional probability distribution that for each value $V$ of *INT_START* sets the probability of all values of *INT_END* that are smaller than $V$ to be zero.

The duration of any particular interval is represented by a variable *INT_DUR* that has a value that is the difference between *INT_START* and *INT_END*. Figure 5.1 demonstrates graphically the representation for an interval and its duration.

### 5.1.2. Building patterns and other temporal abstractions

Utilizing the association of indexed variables with corresponding time points in an MTBN model we can build patterns and other useful abstractions. These abstractions that are created directly from time-indexed variables are **first-order abstractions**. We can build subsequently **higher-order abstractions** by further abstracting over them. For instance, consider figure 5.2. By examining the temporally indexed variable *GLUCOSE*, we can define the first-order abstraction *ELEVATED_GLUCOSE*. By examining the temporally indexed variable *CHOLESTEROL*, we can define the first-order abstraction *ELEVATED_CHOLESTEROL*. We then define a second-order abstraction *ELEVATED_GLUCOSE_AND_CHOLESTEROL*. As long as we define the appropriate conditional probability distributions correctly, there is no limit to the number, degree, and form of abstractions we can create.

### 5.1.3. Facts, events

The way in which facts and events are defined and modeled is application-dependent. For example, we can follow Allen's classic definition of a fact as a property that holds over all subintervals of the interval associated with that fact [Allen 1984]. An event could be defined as a property of an interval that does not hold within all subintervals (i.e., is invalid for some time point(s)).

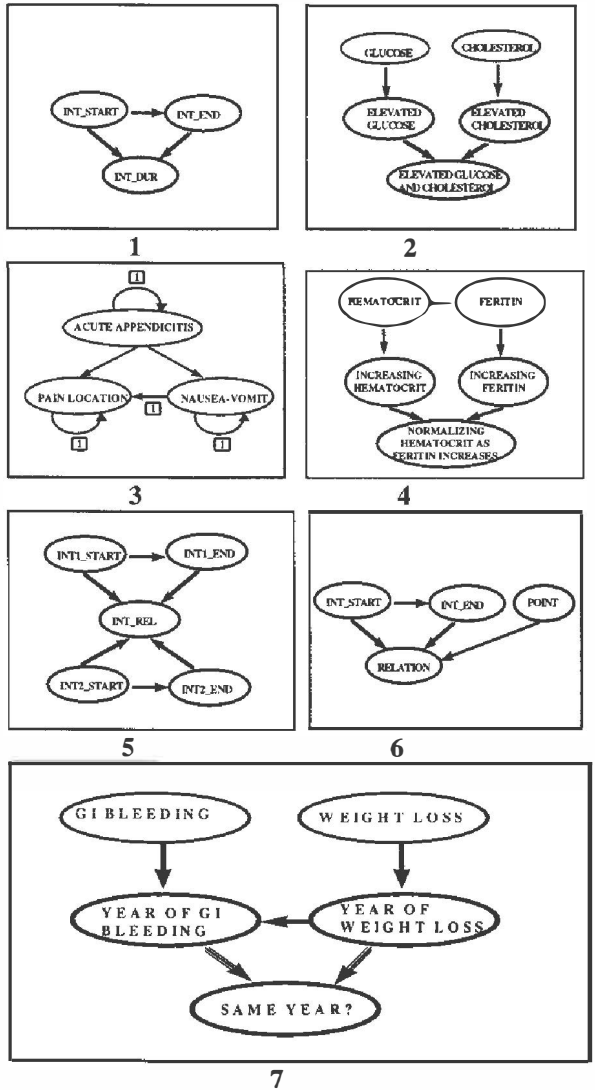

Figure 5: Examples of standard BN time-modeling techniques that apply to MTBNs as well

### 5.1.4. Spatio-temporal reasoning

Figure 5.3 shows a textbook example of the spatio-temporal evolution of a disease process. In particular, *ACUTE_APPENDICITIS* is typically causing pain that starts from the right hypochondrial area and proceeds to the right iliohypogastric area within a few hours. *NAUSEA_VOMIT* are additional findings that accompany this temporal evolution to form a temporal pattern with high diagnostic significance.

### 5.1.5. Reasoning about temporal entities

Time modeling requires not only the ability to represent temporal constructs (entities and relationships) but also to reason about those, as well as reasoning about time itself. In figure 5.4 we model the relationship of two patterns of indexed lab measurements. In this example we represent the composite fact that the *HEMATOCRIT* level in a patient is returning to normal as *FERITIN* levels are increasing. In figure 5.5 we present an example of examining the temporal order of two intervals. This is accomplished by connecting the endpoint variables of each interval with a variable *INT_REL*, which takes the following values: {before, after, coincides, meets, follows, overlaps, is_overlapped_by}. In figure 5.6 we present a similar example modeling the relationship of a time point to an interval.

### 5.1.6. Multiple reference to same entities

In an MTBN model we can have multiple representations of the same entity. Figure 5.7 shows the same year can be represented by two different variables (*YEAR_OF_GI_BLEEDING, YEAR_OF_WEIGHT_LOSS*). We can then examine these two to determine whether GI bleeding and weight loss occurred during the same time period (variable *SAME_YEAR?*).

## 5.2. MTBN-SPECIFIC MODELING TECHNIQUES

### 5.2.1. Static and dynamic processes

In MTBNs, processes are defined as composite mechanisms that produce the values of variables in time. As such, an MTBN graph depicts a process as a system of interrelated components associated with temporal probabilistic dependencies and independencies. MTBNs facilitate explicit representation of how a process might evolve over time. This is because individual mechanisms within the process are activated or not at different time points, and the time difference (lag) between causes and effects is explicitly defined as a random variable.

### 5.2.2. Persistence

An important property of many variables is that they tend to persist over time. That is, their values are not only determined by external factors but by their previous values. Figure 6.1 shows an example where variable *CREATININE* is not only affected by the variable *RENAL_FUNCTION*, but by *CREATININE* at the previous time point as well.

### 5.2.3. Feedback loops

Many processes in biology, medicine, economics, and other fields, exhibit directed causal loops in which one variable determines another one, which in turn determines the first variable and so on. Such loops give rise to distributions that are extremely difficult to represent with simple BNs, unless an equilibrium state has been reached. In MTBN-SGs, we are not restricted to modeling stable feedback processes only. We can model any part of the feedback process, even if no equilibrium *has or can* be reached, by examining the loop over time. The behavior of the feedback system emerges as a result of local interactions defined by the knowledge engineer. It is in the engineer's discretion to decide between stable and unstable feedback modeling. Figure 1 shows an example of the use of MTBNs to model feedback. There *GLUCOSE* increases the secretion of *INSULIN* from the pancreas, and *INSULIN* causes *GLUCOSE* to drop due to absorption from the body cells.

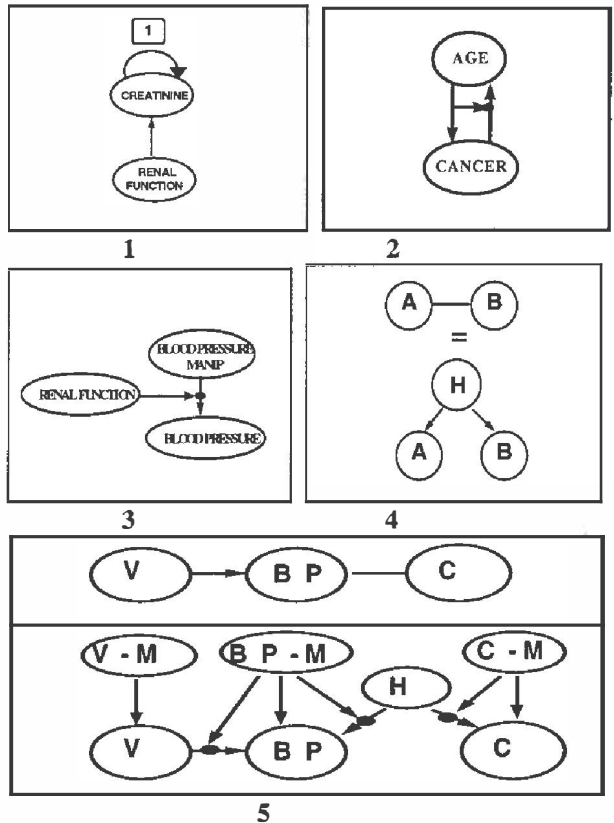

Figure 6: Examples of time-modeling techniques that are facilitated by MTBNs

### 5.2.4. Types of uncertainty and mixtures of causal processes

Several types of uncertainty can be represented in MTBNs. We already showed that **causal uncertainty** is represented in the form of a probabilistic causal arc from cause to effect. **Unmodeled (i.e., exogenous) uncertainty** is represented as a prior distribution of the nodes that have no parents. Finally, **interval uncertainty** is modeled with the use of random-variable interval endpoints (section 5.1.1.).

Another important technique is when we model uncertainty about the causal structure of the process. In the example of figure 6.2, we are uncertain whether variable *AGE* is causing *CANCER* (possible explanation: diminished capacity of the immune system to dispose of genetically damaged cells), or whether *CANCER* is causing *AGE* (possible explanation: not having cancer causes higher survival and thus observed age in a given group). We can model both processes in the same model and quantify our belief for each of the possible alternatives. An arc among the arcs can be used to force them to be (concurrently) mutually exclusive. The same technique can be used to represent reciprocal causation, and structural uncertainty *in a temporal context.* That is, we can represent **mixtures of distributions** that are produced by causally incompatible BN processes using a *single* MTBN.

### 5.2.5. Flexible actions/decisions and decision-theoretic models

Due to the similarities of MTBNs to BNs and influence diagrams (IDs), MTBNs can be readily extended to model preferences (utilities) and to perform decision-theoretic inference. MTBNs do not restrict us to incorporate a model of the decision-maker to the domain model. This gives the flexibility to go beyond traditional ID conventions like the *non-forgetting decision maker, the single decision maker, and free will principles* [Howard et al. 1981]. In effect, with MTBNs, it is straightforward to model restrictions on the availability of decision options according to domain events, multiple decision makers, and information loss. To keep the model's definition and properties as simple as possible, we decided to *not* give special semantic status to actions/decisions. Instead we utilize appropriate conventions at the implementation level.

In figure 6.3 we show an example of a modeling convention that indicates which variables in the model are to be potentially manipulated. Variable *BLOOD_ PRESSURE* is such an example, and for it there is a specially marked variable called *BLOOD_ PRESSURE_ MANIP* that has a deterministic relationship with *BLOOD_PRESSURE* (implemented via a conditional probability distribution that assigns to *BLOOD_ PRESSURE* exactly the value of *BLOOD_ PRESSURE_ MANIP*). The MTBN manipulation convention consists of three rules:

(a) Causal manipulations of variables are allowed by an inference program only through such manipulation dummy variables.
(b) When a manipulation variable is instantiated, every other arc (besides its own) to the children of that variable is ignored (deactivated).
(c) Otherwise the manipulation variable is ignored.

In figure 6.3, we also show an example where a domain variable determines which alternative manipulation options we have with respect to a decision variable. In the example, *RENAL_FUNCTION* restricts the ability to manipulate *BLOOD_PRESSURE* in a patient.

**5.2.6. Mixture of causal and non-causal associations**

An important modeling problem arises when we want to represent mixtures of causal and non-causal (i.e., associational) relationships and manipulation variables (i.e., decisions) simultaneously. This situation can happen in both temporal and atemporal problem-solving contexts. The nature of the problem is that we want the effects of our variable manipulations to be propagated along causal paths only. As in the case of causal manipulations, we maintain causal semantics for MTBNs, and implement non-causal associations (as well as mixtures of both) utilizing introduced variables.

In figure 6.4 we present a simple transformation of the original association between $A$ and $B$ to produce a causal graph with a hidden variable $H$ being introduced. First note that in general we focus on non-causal associations that are created by one or more common causes. We do not attempt at this point to model non-causal associations that are created by selection bias. The transformation consists of the following rules:

(a) For all non-causal arcs between variables $V_i$ and $V_j$, substitute each non-causal arc with a pair of causal arcs emanating from a unique (for the pair) uninstantiated dummy variable $H_k$ which is not connected to any other variable (except the two nodes $V_i$, $V_j$), and has a conditional probability distribution with each child that captures exactly the original relationship between $V_i$ and $V_j$ (Note: if the dependencies and independencies of the resulting graph are not the same as the original, this means that the original was not consistent with our interpretation of non-causal association, and thus has to be modified).

(b) Apply the causal manipulation convention of subsection 5.2.5, with the addition of conditioning any incoming arc to the manipulated variable $Z$ from a dummy variable so that a manipulation of $Z$ deactivates such arcs.

The restriction to paired hidden variables and the absence of selection bias means that the ability to model noncausal associations is currently limited. These restrictions were made for pragmatic reasons. It may be possible to relax them and still maintain computational tractability.

In figure 6.5 we use this convention to create an example of a mixture of causal and non-causal associations. The upper part of the figure represents this mixture schematically. The lower part is the converted MTBN that captures the required relationships. More specifically, the variable *Vasodilator (V)* causally influences *Blood_pressure (BP)*, but *Blood_pressure* is not causally associated with *Cataract (C)*. This means that although *Blood_pressure* determines stochastically *Cataract* and *Cataract* is dependent on *Vasodilator* given *Blood_pressure* (according to the Markov and Faithfulness conditions [Spirtes et al. 1993]), if we causally manipulate *Cataract* (variable *C-M*) this is not going to change *Blood_pressure*, nor change our belief about *Vasodilator*. On the other hand, if we observe *Cataract*, this will change our belief about (the unobserved) *Blood_pressure*. In comparison, both observing and manipulating *Vasodilator* (variable *V-M*) will give information about *Blood_pressure*. Finally, if we observe *Blood_pressure*, then *Vasodilator* should be dependent on *Cataract*, but if we manipulate *Blood_pressure* (variable *BP-M*), then *Vasodilator* should be independent of *Cataract*.

**6. COMPARISON OF MTBN-SGs AND OTHER REPRESENTATIONS**

Since MTBN-SGs extend BNs structurally as well as temporally, it is worthwhile to separate a class of MTBN-SGs that employs structural extensions only. In particular we will define MBNs as *the class of MTBN-SGs with no temporally-indexed variables.* The example MTBN of figure 6.2 is an MBN. MBNs differ from BNs in that (i) they allow arcs between arcs and (ii) arcs can have a prior distribution. It is also evident that BNs are a subclass of MBNs, since BNs are the class of MBNs with all arcs instantiated to the value "active" with probability 1.

Table 2 : Differences among MTBN classes and DBNs.

|  | BN | MBN | DBNs | MTBN-SG |
|---|---|---|---|---|
| TIME-STAMPED VARIABLES | NO | NO | YES | YES |
| DYNAMIC STRUCTURE | NO | YES (ATEMPORAL) | NO | YES (TEMPORAL) |
| EXPLICIT TIME | NO | NO | YES | YES |
| MORE THAN ONE ABSTRACTION LEVELS | NO | NO | NO | YES (1 TEMPORAL GRANULARITY AND ABSTRACTED VARIABLES) |
| TARGETED TASKS | GENERAL PURPOSE | GENERAL PURPOSE | ADAPTIVE FORECASTING AND CONTROL | GENERAL PURPOSE |

Since these three formalisms are different in representational complexity, established algorithms, commercial implementations, etc., the discrimination among them allows us to suggest the one that may be more appropriate for a particular task. Table 2 summarizes the differences of the 3 formalisms. In the table we also include standard dynamic BNs (DBNs), which is a formal temporal extension of BNs, as a point of reference. We note that DBNs could be viewed as a subclass of MTBN-SGs in which there are prototypical conditional probability functions, constant (i.e., not explicitly dynamic) mechanisms and lags, and all variables are temporally indexed.

By looking at the table, we can see that BNs are more appropriate for tasks with a single non-dynamic process. MBNs are more suitable for mixtures of structurally different atemporal processes (i.e., processes that can not be causally represented by a single BN structure), as well as for cases where it is important to explicitly denote that certain variables in the model restrict or enable the causal influences among other variables (see sections 5.2.5, and 5.2.6 for examples). DBNs do have explicit temporal semantics, but only one level of temporal granularity and no abstraction. DBNs support special relationships among variables, designed to increase efficiency by restricting the class of relationships that are allowed to be represented. DBNs are especially appropriate for multivariate Bayesian time-series analysis applications where a qualitative structure can be learned from an expert and the appropriate parameterization from data, due to the development of techniques for adaptation of the network to new data instances. MTBN-SGs are appropriate for tasks that involve temporally explicit representations, where explicit time modeling is required, and where mixing explicit time with abstracted time modeling can lead to easier knowledge/data acquisition as well as greater efficiency (due to smaller and simpler models).

## 7. DISCUSSION

Several extensions of BNs for time modeling have been presented over the last few years. These include temporal influence diagrams [Provan 1993], dynamic BNs [Dagum and Galper 1993A], temporal models of endogenous change [Hanks et al. 1995], action networks [Darwiche et al. 1994], embedded Markov processes [Berzuini et al. 1992], logic and time nets [Kanazawa 1991], knowledge-based model construction methods from temporal logics [Ngo et al. 1995, Glesner et al. 1995], as well as specific applications [Polaschek et al. 1993, Berzuini et al. 1992]. Although many BN variants have been introduced for time and causal modeling, many have not provided formal semantics for the models, nor have they dealt with the interaction of causal with temporal semantics. The lack of such concerns can lead to models that violate fundamental principles, such as that effects can not precede their causes [Drudzel et al. 1993].

In this paper we introduced MTBNs and provided a well-defined causal and temporal interpretation. MTBN-SGs have some unique features relative to other temporal BN variants. They have explicitly modeled dynamic arc variables, and explicitly modeled lags between causes and effects. They enforce a causal semantics on which non-causal models are built. They utilize a condensed graphical representation for defining and presenting a model, and a deployed form for inference. MTBN-SGs also facilitate the co-existence of temporally explicit (i.e., indexed) with implicit (i.e., abstracted - nonindexed variables). This facilitates knowledge acquisition and computational tractability, while maintaining the explicit temporal semantics. In [Aliferis et al 1995] we show an example of how explicit temporal modeling can lead to completely intractable inference, even for models containing a small number of variables, and how hybrid abstraction modeling can render inference tractable (at the expense of domain query completeness rather than accuracy). Moreover, we emphasize that inference algorithms for MTBNs can utilize the explicit temporal semantics of MTBNs to validate the temporal and causal validity of models expressed in MTBN form.

As in any modeling language, it is the match between the language's features and the domain, as well as the alternative formalisms' features that will determine the appropriateness of that particular language for a problem domain. In the present paper, we provided definitions and basic properties for the MTBN-SG modeling language. We also discussed modeling techniques that are facilitated by MTBNs, as well as modeling that is inherited from their affinity to BNs. Weaknesses of MTBNs are that they do not suggest or enforce any general theory of time for decision-support systems, nor do they come with a precise specification of a universal temporal ontology. More importantly, their expressive power is limited by their propositional nature and restricted description of properties of individuals or groups of individual entities. Thus, it is very important, as with every modeling tool or methodology, to select the application domain after careful consideration of the representation's strengths and weaknesses. We believe that MTBNs are excellent candidates for many normative uncertain temporal reasoning tasks that involve prediction, diagnosis, and optimal decision selection in complex dynamic domains.

Our current work focuses on (a) implementing multiple granularity extensions to MTBNs, (b) increasing the flexibility of defining classes of model components and instantiating them as needed to solve particular queries, and (c) comparing MTBNs against BNs in solving real-life complex temporal diagnostic and policy formulation problems in the domain of liver transplantation.

## Acknowledgments

This work has significantly benefited from suggestions of Drs Bruce Buchanan, Martha Pollack, and Michael Wagner. We are thankful to Dr Eric Horvitz and the anonymous reviewers for many useful recommendations. Support from the National Science Foundation (grant BES-9315428), and the National Library of Medicine (grant LM05291-02) is gratefully acknowledged.

## References

C.F. Aliferis, G.F. Cooper "A New Formalism for Temporal Modeling in Medical Decision-Support Systems" Proceedings of the $18^{th}$ Annual Symposium on Computer Applications in Medical Care, 1995, p. 213-217.

C.F. Aliferis, G.F. Cooper "Modifiable Temporal Belief Networks With a Single Temporal Granularity", Technical Report 96-01, Section of Medical Informatics, University of Pittsburgh, 1996.

J.F. Allen "Towards a General Theory of Action and Time" Artificial Intelligence 23(2):123-154, 1984.

C. Berzuini et al "Bayesian Networks for Patient Monitoring" Artificial Intelligence in Medicine 4:243-260, 1992.

R. Bouckaert "Properties of Bayesian Belief Network Learning Algorithms" Proceedings of Uncertainty in Artificial Intelligence, 1994, p. 102-109.

G.F. Cooper "The Computational Complexity of Probabilistic Inference Using Belief Networks" Artificial Intelligence 42:393-405, 1990.

G.F. Cooper, E. Herskovits "A Bayesian Method for the Induction of Probabilistic Networks from Data" Machine Learning, 9:309-347, 1992.

P. Dagum, A. Galper "Forecasting Sleep Apnea with Dynamic Network Models" Proceedings of Uncertainty in Artificial Intelligence, 1993, p. 64-71. [A]

P. Dagum, M.Luby "Approximating Probabilistic Inference in Bayesian Belief Networks is NP-hard" Artificial Intelligence 60:141-153, 1993. [B]

A. Darwiche, M. Goldszmidt "Action Networks: A Framework for Reasoning about Actions and Change under Uncertainty", Proceedings of Uncertainty in Artificial Intelligence, 1994, p. 136-144.

M. Drudzel, H. Simon "Causality in Bayesian Belief Networks" Proceedings of Uncertainty in Artificial Intelligence, 1993, p. 3-11.

B.C. van Fraassen "An introduction to the Philosophy of Time and Space" Columbia university Press, New York, 1970.

S. Glesner, D. Koller "Constructing Flexible Dynamic Belief Networks from First-Order Probabilistic Knowledge Bases", in Lecture Notes in Artificial Intelligence, C. Froidevaux and J. Kohlas (Eds.) p. 217-226, Springer Verlag, 1995.

P. Haddaway "Representing Plans Under Uncertainty: A Logic of Time, Chance, and Action" vol. 770 of Lecture Notes in Artificial Intelligence, Seattle, 1994.

S. Hanks, D. Madigan, J. Gavrin "Probabilistic Temporal Reasoning with Endogenous Change" Proceedings of Uncertainty in Artificial Intelligence, 1995, p. 245-254.

D. Heckerman, D. Geiger, D.M. Chikering "Learning Bayesian Networks: The Combination of Knowledge and Statistical Data" Proceedings of Uncertainty in Artificial Intelligence, 1994, 293-301.

R.A. Howard, J.E. Matheson "Influence diagrams" in: "The Principles and Applications of Decision Analysis" Strategic Decisions Group, Menlo Park California, 1981.

K. Kanazawa "A Logic and Time Nets for Probabilistic Inference" " Proceedings of National Conference of the American Association for Artificial Intelligence, 1991, 360-365.

R.E. Neapolitan "Probabilistic Reasoning in Expert Systems" John Wiley and Sons, New York, 1990.

L. Ngo, P. Haddaway, J. Helwig "A Theoretical Framework for Context-Sensitive Temporal Probability Model Construction with Application to Plan Projection", Proceedings of Uncertainty in Artificial Intelligence, 1995, 419-426.

J. Pearl "Probabilistic Reasoning in Intelligent Systems" Morgan Kafmann, San Mateo, California, 1988

J.X. Polaschek et al. "Using belief networks to interpret qualitative data in the ICU" Respiratory Care 38 (1):60-71, 1993.

G. Provan "Tradeoffs in Constructing and Evaluating Temporal Influence Diagrams" Proceedings of Uncertainty in Artificial Intelligence, 1993, 40-47.

P. Spirtes, C. Glymour, R. Scheines "Causation, Prediction and Search" Springer Verlag, New York, 1992.

HJ Suermondt, GF Cooper "An Evaluation of Probabilistic Inference" Comp Biomed Res 26;242-254, 1993.

A.U.Tansel et al. "Temporal Databases", The Benjamin/Cummings Publishing Company, Redwood city California, 1993.